\documentclass[twocolumn]{IEEEtran}
\usepackage{lscape}
\usepackage{booktabs}
\usepackage{authblk}
\usepackage{url}
\usepackage{silence}
\usepackage[font=scriptsize]{caption}

\usepackage[pdftex]{graphicx}
\usepackage{epstopdf}
\usepackage{pifont}
\usepackage{array}
\usepackage{physics}
\usepackage{verse}
\usepackage{longtable}
\usepackage{supertabular}
\usepackage{cite}
\usepackage{color}
\usepackage{amsmath}
\usepackage{lettrine}
\usepackage{hyperref}
\usepackage{multirow}
\usepackage{multicol}
\usepackage{dirtytalk}
\usepackage{color}
\usepackage{caption}
\usepackage{subcaption}

\begin{document}

\title{R\textsuperscript{2}S100K: Road-Region Segmentation Dataset For Semi-Supervised Autonomous Driving in the Wild}

\author{Muhammad Atif Butt$^{1,2}$, Hassan Ali$^1$, Adnan Qayyum$^1$, Waqas Sultani$^1$, Ala Al-Fuqaha$^{3}$, and Junaid Qadir$^{4*}$\\
$^1$ Information Technology University (ITU), Punjab, Lahore, Pakistan.\\
$^2$ Computer Vision Center, Universitat Autonoma Barcelona, Spain.\\
$^3$ Information and Computing Technology (ITC) Division, College of Science and Engineering, \\ Hamad Bin Khalifa University, Doha, Qatar. \\
$^4$ Qatar University, Doha, Qatar.\\
%(\textit{Corresponding author: Junaid Qadir (junaid.qadir@itu.edu.pk)})
\textit{$^*$Corresponding author: Junaid Qadir (jqadir@qu.edu.qa) }
}

\maketitle

\begin{abstract}
\label{abstract}
Semantic understanding of roadways is a key enabling factor for safe autonomous driving. However, existing autonomous driving datasets provide well-structured urban roads while ignoring unstructured roadways containing distress, potholes, water puddles, and various kinds of road patches i.e., earthen, gravel etc. To this end, we introduce \textbf{R}oad \textbf{R}egion \textbf{S}egmentation dataset (R\textsuperscript{2}S100K)---a large-scale dataset and benchmark for training and evaluation of road segmentation in aforementioned challenging {unstructured} roadways. R\textsuperscript{2}S100K comprises 100K images extracted from a large and diverse set of video sequences covering more than 1000 KM of roadways. Out of these 100K privacy respecting images, 14,000 images have fine pixel-labeling of road regions, with 86,000 unlabeled images that can be leveraged through semi-supervised learning methods. Alongside, we present an \textbf{E}fficient \textbf{D}ata \textbf{S}ampling (EDS) based self-training framework to improve learning by leveraging unlabeled data. Our experimental results demonstrate that the proposed method significantly improves learning methods in generalizability and reduces the labeling cost for semantic segmentation tasks. Our benchmark will be publicly available to facilitate future research at \url{https://r2s100k.github.io/}.
\end{abstract}

\section{Introduction}
\label{sec:intro}
\begin{figure*}[t]
\centering
\includegraphics[width=\textwidth]{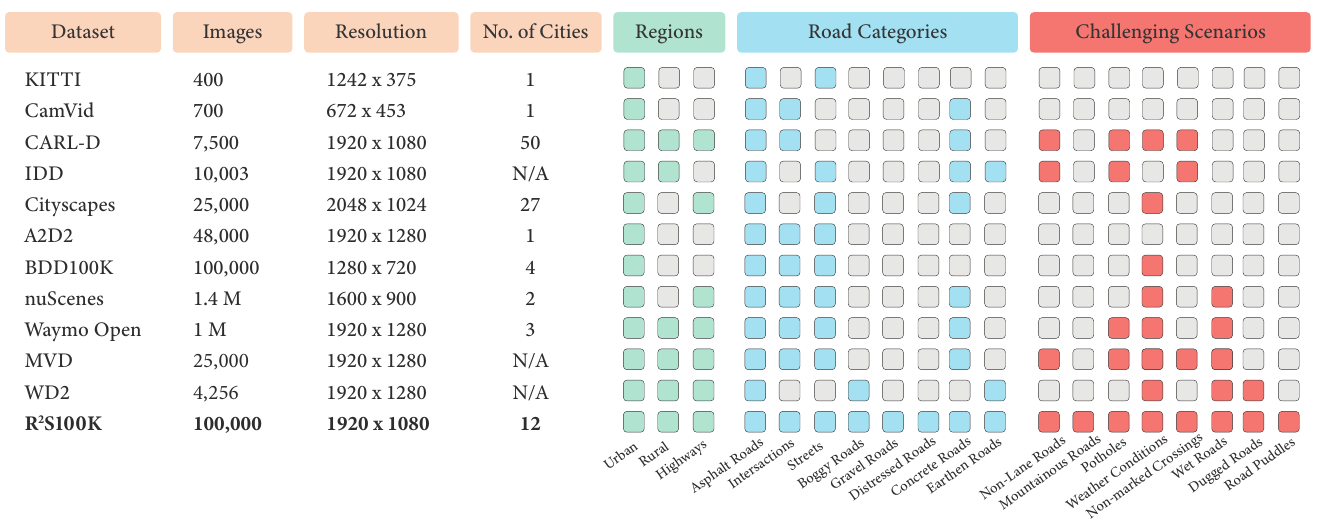}
\caption{Comparison of dataset statistics with existing driving datasets i.e., KITTI~\cite{geiger2012we}, CamVid~\cite{brostow2009semantic}, CARL-D~\cite{butt2022carl}, IDD~\cite{varma2019idd}, Cityscapes~\cite{cordts2016cityscapes}, A2D2~\cite{geyer2020a2d2}, BDD100K~\cite{yu2020bdd100k}, nuScenes~\cite{caesar2020nuscenes}, Waymo~\cite{sun2020scalability}, MVD~\cite{neuhold2017mapillary}, and Wilddash~\cite{zendel2018wilddash}. \textit{Our R\textsuperscript{2}S100K covers more diverse road infrastructure and challenging scenarios as compared to the existing benchmarks. Therefore, our dataset can be used in developing more robust and generalized road segmentation methods for autonomous driving.}}
\label{fig:datasets}
\end{figure*}

Visual perception for recognizing objects, obstacles, and pedestrians is a core building block for efficient autonomous driving. Recently, semantic segmentation has emerged as an efficient perception method that aims to determine the semantic labels for each pixel of an image \cite{siam2018comparative}. Thanks to the availability of rich scene segmentation datasets (discussed in Figure \ref{fig:datasets}), significant technical progress has been made in this direction. However, several formidable challenges still remain on the path to efficient autonomous driving in the wild.

% \begin{figure}[t]
% \begin{center}
% %\fbox{\rule{0pt}{2in} \rule{0.9\linewidth}{0pt}}
%     \includegraphics[width=\linewidth]{Figures/Dataset-Images-Col.jpg}
% \end{center}
% \caption{Examples of challenging roadways in our dataset. \textit{Instead of considering whole paved road region as one class, we distinguish safe asphalt \testclr{green!40!white} road region and its associated atypical classes found on unstructured roads such as distress \testclr{violet!60} , wet surface \testclr{OliveGreen} , gravel \testclr{gray!60} , boggy \testclr{brown} , vegetation misc. \testclr{LimeGreen} , crag-stone \testclr{black!60!Tan!90} , road grime \testclr{brown!70} , drainage grate \testclr{red!37!green!34!blue!65} , earthen \testclr{Apricot} , water puddle \testclr{Periwinkle!60} , misc. \testclr{BrickRed!80} , speed breakers \testclr{BurntOrange} , and concrete \testclr{Maroon} road patches.}}
% \label{fig:dataset-samples}
% \end{figure}

\begin{figure*}[t]
\begin{center}
    \includegraphics[width=\linewidth]{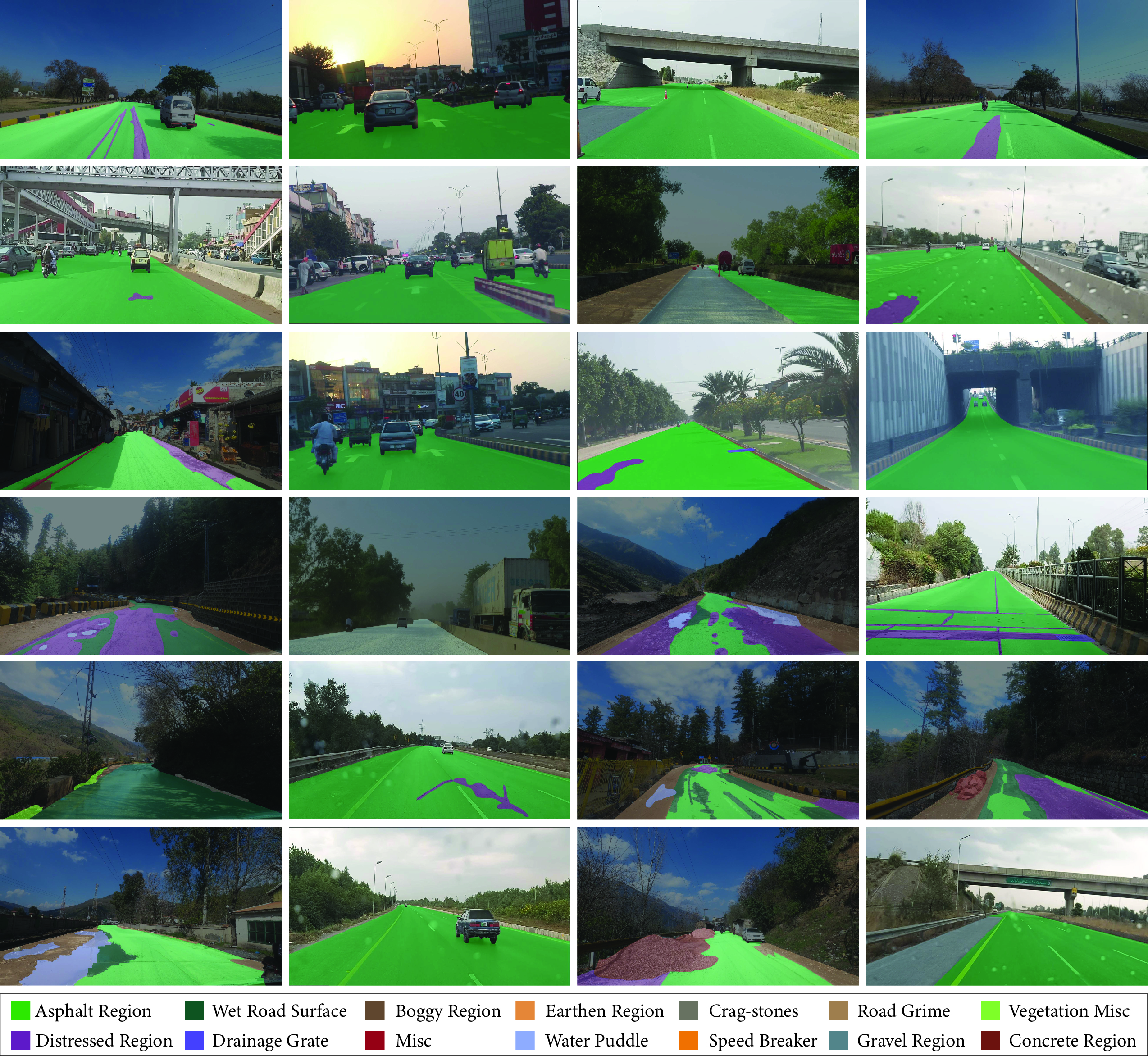}
\end{center}
\caption{Examples of our dataset images covering a wide array of roadways, varying across different lighting and weather conditions. \textit{Instead of considering the whole paved road region as one class, we distinguish safe asphalt road region and its associated atypical classes found on unstructured roads such as distress, wet surface, gravel, boggy, vegetation misc., crag-stone, road grime, drainage grate, earthen, water puddle, misc., speed breakers, and concrete road patches.}}
\label{fig:dataset-samples}
\end{figure*}

\textit{Firstly}, existing autonomous driving datasets \cite{geiger2012we, brostow2009semantic, cordts2016cityscapes, yu2020bdd100k, caesar2020nuscenes, sun2020scalability} are not generalized; they cover well-paved urban roads of developed countries which represents 3.7\% road infrastructure of the world~\cite{Roads} and barely serve 17\% of the total world's population~\cite{population}. More recently, Segment Anything~\cite{kirillov2023segment}---the largest segmentation dataset with more than one billion masks for 11 million images has been released to perform general purpose segmentation tasks. However, despite being largest in size, it only covers 0.9\% of data samples from low-income countries. Therefore, these datasets have scant coverage of unstructured roadways containing hazardous road patches (i.e., distress, earthen, gravel) that are common in developing world, as shown in Figure~\ref{fig:dataset-samples}. The presence of such ambiguous road regions poses an enormous hazard to human drivers and lead towards severe road accidents and fatalities. According to World Health Organization (WHO), 1.3 million people die every year due to road accidents~\cite{world2020world} with 93\% of causalities occurring in low- and middle-income countries. The global road safety report points out that non-standard road infrastructure is a key reason for higher road accident rates in these countries~\cite{world2019world}. Therefore, under representation of such challenging data in existing datasets is a critical omission for research on autonomous driving and an indication towards the need of a benchmark to improve autonomous driving in such challenging road scenarios.

\textit{Secondly}, pixel-level annotation of images is excessively expensive---for cityscapes, labeling an image took an hour on average \cite{cordts2016cityscapes}---leading to smaller segmentation datasets than in other domains \cite{lin2014microsoft,deng2009imagenet}, consequently limiting the generalizability of the trained models. Although semi-supervised learning methods \cite{abdalla2019fine, he2019knowledge, hung2018adversarial, yu2022self} have been proposed that leverage unlabeled data to improve learning, these methods suffer limitations because: (i) segmentation datasets are often highly imbalanced in terms of pixel counts corresponding to each class \cite{rezaei2020recurrent}, and different physical scenarios in which the dataset is collected. Therefore, the resulting model performs significantly worse in physical scenarios that are not common (e.g. rare weather conditions and unstructured roads), which can be lethal in autonomous driving; (ii) Biased predictions caused by the data imbalance in early semi-supervised training phase \cite{he2019knowledge} lead to a higher misclassification rate during inference; (iii) self-training segmentation models is computationally very expensive due to a large number of pseudo labels \cite{wei2018revisiting}. In this regard, there is a need of an efficient method to improve performance while considering accuracy-energy trade-offs. To address these challenges, we have made the following contributions:

\begin{enumerate}
    \item We introduce \textbf{R}oad \textbf{R}egion \textbf{S}egmentation (R\textsuperscript{2}S100K) dataset for autonomous driving comprising 100K diverse set of road images, covering 1000+ KMs of challenging roadways, as shown in Figure~\ref{fig:dataset-samples}. R\textsuperscript{2}S100K dataset covers more challenging road categories and scenarios as compared to existing datasets. Moreover, R2S100K serves as an initial step in representing unstructured roads prevalent in low-income countries, allowing for a more comprehensive stress-testing of foundational segmentation models for autonomous driving. 

    \item We propose an unsupervised \textbf{E}fficient \textbf{D}ata \textbf{S}ampling (EDS) method to sample a subset from the unlabelled training data, which offers three benefits: (i) EDS notably alleviates the data imbalance in the physical scenarios, (ii) improves the performance of supervised (0.71\%-6.72\% MIoU) and semi-supervised (0.26\%-1.84\% MIoU) models, and (iii) significantly reduces the annotation and training costs (75\% fewer pseudo-labels and 79\% decrease in the training time).
    
     \item The EDS is compatible with multiple learning frameworks (supervised, semi-supervised), model architectures, and  can be integrated with existing datasets such as Cityscapes, CamVid, and BDD100K due to similar labeling schema.
\end{enumerate}

\section{Background}
\label{sec:back}
\subsection{Autonomous Driving Datasets}
In the past couple of years, several datasets have been released to accelerate the development of visual perception algorithms. These datasets can be categorized into two major groups: (i) object detection---which focuses on 2D/3D objects~\cite{geiger2012we,sun2020scalability,caesar2020nuscenes,huang2019apolloscape,xiao2021pandaset,dollar2009pedestrian, zhang2017citypersons}; and (ii) scene segmentation---which focuses on semantic segmentation for scene understanding. Here we discuss some important characteristics of these datasets.

\textbf{Object Detection Datasets:} KITTI~\cite{geiger2012we} is one of the most widely used vision benchmark suites for object detection on urban roads and highways which contains 15k images along with 200k annotations. Later on, Waymo open dataset~\cite{sun2020scalability} presented more than 23 million 2D and 3D bounding boxes annotations of 1,150 inter-cities urban scene segments. nuScenes~\cite{caesar2020nuscenes} presented 1.4 million 3D bounding box annotations of 1000 urban as well as suburban road scenes for 23 classes. In 2019, ApolloScape dataset~\cite{huang2019apolloscape} has been released with comprises 70k 3D annotations along with 160k semantic mask annotations of urban roads and highways under varying weather conditions. Similarly, Pandaset~\cite{xiao2021pandaset} presented 1 million 3D bounding box annotations for object detection in urban traffic scenarios. Other than these a few, various datasets \cite{dollar2009pedestrian, caesar2020nuscenes, zhang2017citypersons} have been proposed which played an important role in developing efficient object detection and recognition algorithms.

\textbf{Semantic Segmentation Datasets:}
CamVid~\cite{brostow2009semantic} is considered among the pioneer scene segmentation datasets---comprising 700 fine annotations for 32 classes. In 2016, Cityscapes~\cite{cordts2016cityscapes} is released which contains 5000 fine and 20,000 coarse-annotations for urban roads. In 2017, Mapillary Dataset~\cite{neuhold2017mapillary} comprising 25K fine annotations of inter-continental urban scenes was presented. Later on, BDD100K~\cite{yu2020bdd100k} is released in 2020 which provides 10K fine annotations of urban roadways. MVD~\cite{neuhold2017mapillary} contains 25K images covering diverse yet urban roadways. 

Though, these datasets provide enriched information of urban scenes for scene segmentation tasks, however, they do not cover unstructured road conditions and hazardous road patches which are commonly encountered in developing countries. Therefore, models trained on these datasets cannot be generalized to the challenging roadways. Other than urban driving, a few datasets have been released for visual perception in off-road driving scenarios. OFFSEG~\cite{viswanath2021offseg} framework covers RELLIS-3D~\cite{jiang2021rellis} containing 6,235 images, and RUGD~\cite{wigness2019rugd} comprising of 7546 images of outdoor off-road driving scenes. Wilddash-v2 contains 4256 images~\cite{zendel2018wilddash} covers unstructured road classes like distress, and gravel patches. However, they label these classes under single \textit{Road} class rather than distinguishing as safe and hazardous regions. Recently, CARL-D~\cite{butt2022carl,rasib2021pixel}, and IDD~\cite{varma2019idd} datasets have also been released which provide annotations of urban and rural roads, however, they still lack aforementioned hazardous road patches that can highly influence the performance of autonomous driving models.

\begin{figure*}[t]
\begin{center}
    \includegraphics[width=\linewidth]{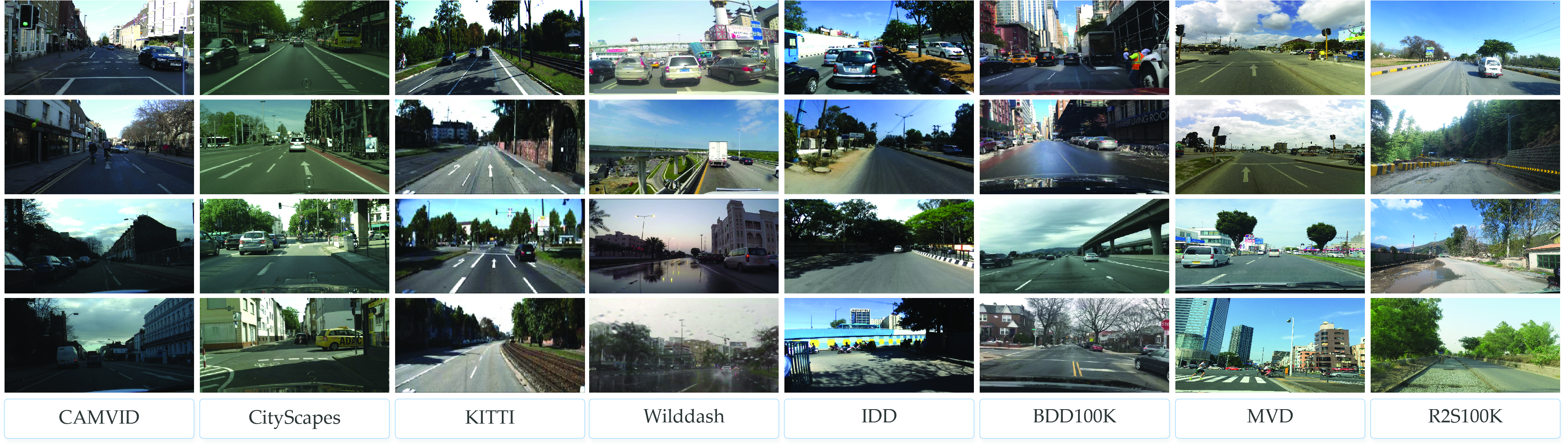}
\end{center}
\caption{Examples of road types covered in existing autonomous driving datasets for visual scene segmentation. \textit{R\textsuperscript{2}S100K covers more challenging/hazardous roads in both---the rural and rural areas. While, most of the existing datasets focus on the well-paved road infrastructure of urban areas, and do not distinguish among safe and hazardous road regions.}}
\label{fig:datasets-comparison}
\end{figure*}

\subsection{Scene Segmentation Methods}
\textbf{Fully Supervised Learning}: Since the pioneering work of FCN \cite{long2015fully}, significant progress has been made in developing more deeper neural networks for semantic segmentation tasks. The semantic segmentation model aims to predict the semantic category of each pixel from a given label set and segment the input image according to semantic information---suggested by Long et al. \cite{long2015fully}. The FCN outperforms conventional approaches by 20\% on Pascal VOC dataset. The U-net is an idea put out by Ronneberger et al. \cite{ronneberger2015u} for segmenting biological images. U-net has a spatial path to maintain spatial information and a context path to learn context knowledge.

Later on, various supervised methods \cite{noh2015learning, badrinarayanan2017segnet, romera2017erfnet, zhao2017pyramid, chen2014semantic, chen2017deeplab, chen2017rethinking, chen2018encoder, zhao2018icnet, yu2018bisenet, zhang2022laanet} are proposed to perform segmentation tasks in an efficient way. However, these methods employ deep CNNs as backbone networks, which require an immense amount of time to annotate large-scale data which limits the model's capacity to adapt and further improve segmentation performance.

\textbf{Semi-supervised Learning}: Recently, semi-supervised learning methods have demonstrated better applicability in several segmentation domains. Leveraging a huge amount of unlabeled data, these methods have achieved state-of-the-art performance on several segmentation tasks. In literature, several techniques such as video label propagation \cite{mustikovela2016can}, \cite{luc2017predicting}, \cite{budvytis2017large}, knowledge distillation \cite{xie2018improving}, \cite{liu2020structured}, adversarial learning \cite{souly2017semi}, \cite{hung2018adversarial}, and consistency regularization \cite{mittal2019semi} are employed to perform semi-supervised segmentation.

\section{Methodology}
\label{sec:method}
In this section, we describe R\textsuperscript{2}S100K dataset along with our proposed efficient self-training method for semantic segmentation tasks. Figure \ref{fig:datasets} demonstrates a comparison of our dataset with existing datasets. In this section, we introduce a benchmark suite for our proposed Road Region Segmentation Dataset (R\textsuperscript{2}S100K). Firstly, we describe R\textsuperscript{2}S100K in terms of the methodology adopted for data collection, frame selection, labeling, and distribution. Secondly, we discuss the categorization of supervised/ semi-supervised learning methods to develop a benchmark suite for our proposed dataset. In the later section, we discuss our proposed EDS enabled teacher-student based efficient self-training approach to solve the data imbalance problem for semantic segmentation tasks.

\subsection{R\texorpdfstring{$^2$}{^2}S100K}

We present a large-scale R\textsuperscript{2}S100K dataset to train and evaluate supervised/semi-supervised methods in challenging road scenarios. Our dataset can be distinguished from existing datasets in the following three major aspects:

\noindent\textbf{\textit{Distribution Shift}:} R\textsuperscript{2}S100K dataset covers unique and undesiring urban and rural road conditions---described in Table~\ref{Table:class-def} which are commonly encountered while driving, especially in developing countries. Whereas, existing datasets such as KITTI~\cite{geiger2012we}, CamVid~\cite{brostow2009semantic}, Cityscapes~\cite{cordts2016cityscapes}, A2D2~\cite{geyer2020a2d2}, MVD~\cite{neuhold2017mapillary}, BDD100K~\cite{yu2020bdd100k}, nuScenes~\cite{caesar2020nuscenes}, Waymo~\cite{sun2020scalability} represent well developed urban roadways, as depicted in Figure~\ref{fig:datasets-comparison}. IDD though covers distressed and muddy road regions, however, it only distinguishes the mud class from the road and covers damaged road patches under one road class. Moreover, OFFSEG~\cite{viswanath2021offseg} The framework primarily covers off-road driving scenes, which significantly differ from unstructured roadways in terms of representation. Similarly, Wildash~\cite{zendel2018wilddash} covers distress, and gravel patches under a single \textit{Road} class rather than distinguishing them as safe and hazardous regions.

\begin{figure*}[t]
\centering
\includegraphics[width=\textwidth]{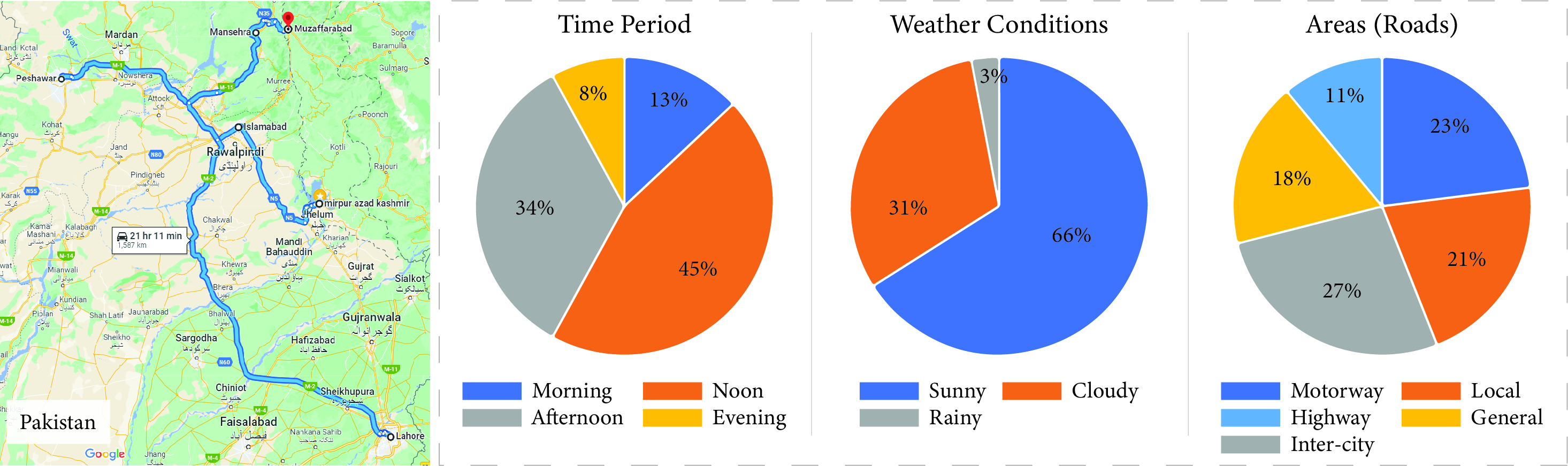}
\caption{Statistical analysis demonstrating the diversity of R\textsuperscript{2}S100K Dataset. (Left) Google Map of route covered for data collection. (Right) Different environmental and infrastructural characteristics: (1) timestamp, (2) weather conditions, and (3) road hierarchy. \textit{We cover over 1000 KMs of roadways of Pakistan---carefully considering the inclusion of motorways, highways, general inter-city and intra-city roads, as well as the rural and hilly areas, under different illuminous and weather conditions.}}
\label{fig:stats}
\end{figure*}

\noindent\textbf{\textit{Diversity}:} R\textsuperscript{2}S100K is constructed over road sequences---captured from 1000+ KMs roadways of Pakistan considering diverse terrain, infrastructural features, and environmental attributes as shown in Figure~\ref{fig:stats}. To ensure diversity in data, we primarily focus on the inclusion of motorways, highways, and urban traffic roads from Punjab, the largest province of Pakistan in terms of population (approximately 127.474 million). Additionally, we extend our coverage to encompass the rural and hilly areas of Khyber-Pakhtunkhwa, the second-largest province by population (approximately 35.53 million), operating under diverse illumination and weather conditions.

\noindent\textbf{\textit{Generalizability}:} R\textsuperscript{2}S100K covers a diverse range of road infrastructure including well-paved asphalt roads along with associated unique hazardous road regions which are categorized as atypical classes, enlisted in Table~\ref{Table:class-def}. However, we assigned distinct labels for our anomalous road classes and used similar labeling schema for asphalt class as cityscapes and BDD100K to ensure the integration of datasets for domain adaptation and semi-supervised learning.

\subsubsection{Data Acquisition}

\paragraph{Driving Platform Setup} A camera is mounted over the dashboard of a standard van with a height of 1.4m from the ground and configured to an aspect ratio of 16:9 to capture the ultimate width of the road. A camera stabilizer is also installed to reduce vibration effects of the vehicle.

%(Right) Absolute pixel count of road classes in existing datasets and our R\textsuperscript{2}S100k. \textit{Alongside diversity, our dataset also contains more labeled road pixels than state-of-the-art datasets.

\noindent\textbf{Road Video Collection:}  
We carefully followed the travel advisory issued by the government to identify diverse roadways. Based on the analysis, we defined a route plan to cover diverse infrastructure for data collection (as shown in Fig. \ref{fig:stats}) to ensure the inclusion of highways, expressways, and general roads of urban cities, rural and hilly areas.

\noindent\textbf{Data Quality Control:} 
We performed pre- and post-collection quality control (QC) to ensure high-quality data collection. In pre-collection QC, the data engineer is required to set up and monitor the data stream of the camera while recording. Whereas, post-collection QC requires data engineers to manually identify and remove the distorted/over-exposed/unclear video sequences.

\noindent\textbf{Data Distribution:}
After data collection under different illumination and weather conditions from 1000+ KM of roadways, distorted/blurry/unclear sequences are excluded, and frames are selected from the remaining video with a 10s difference to avoid redundancy. The vehicle is moving at varying speeds (120 KM/h (motorway), 60-100 KM/h (highway), 20-60 KM/h (within city)). Therefore, the speed variation, blurry sequences exclusion, and 10s difference play a key role in avoiding data redundancy. Lastly, EDS further minimizes the chances of sequential frames in the data. We aligned video sequences to extract the frames to equally distribute the diverse road scenarios. To achieve better diversity, 10 frames are selected after every 10 seconds per frame. Therefore, 100K images of R\textsuperscript{2}S100K dataset are sampled out of 10 million images.

\subsubsection{Data Statistics}

\paragraph{Labeled Data}
The labeled set consists of 14,700 images with fine-layered polygonal annotations which are realized in-house to ensure the highest level of quality. To avoid void spacing and erroneous class overlapping, images are labeled in back to front manner so that no class boundary is dual-labeled. Due to the diversity in data, we categorized road regions into 14 distinct classes as described in Table~\ref{Table:class-def}. 

\paragraph{Unlabeled Data}
The unlabeled set of our dataset contains 86,000 images, covering diverse road infrastructure. As shown in Figure \ref{fig:stats}, our unlabeled set is collected under varying weather conditions and time periods to ensure diversity in terms of downstream autonomous driving tasks.

\begin{figure}[t]
    \centering
    \fbox{\includegraphics[width=\linewidth]{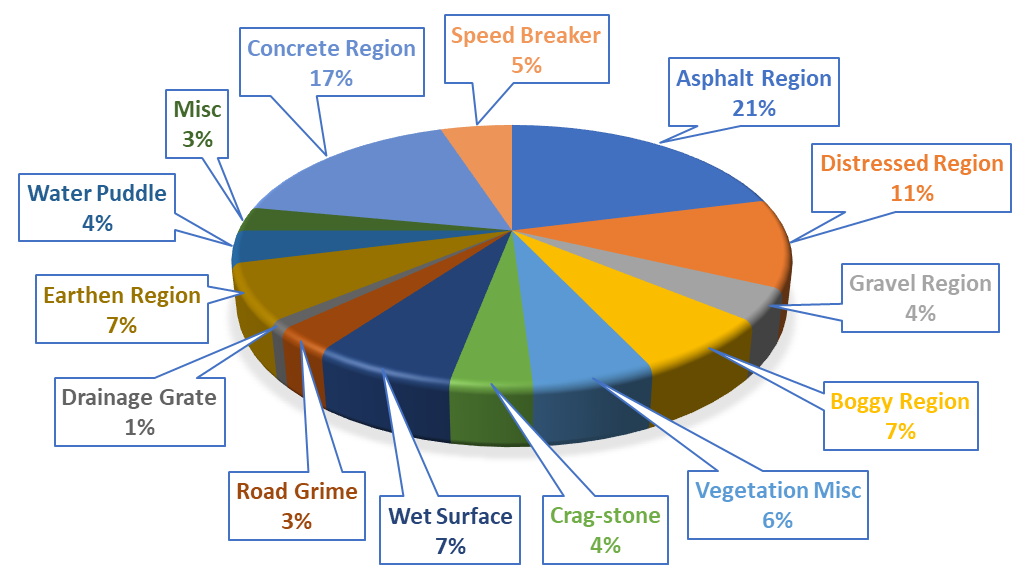}}
    \caption{Distribution of road classes in R\textsuperscript{2}S100K. \textit{Asphalt and concrete regions represent the safe drivable road regions with the higher representation among the other hazardous road patches.}}
    \label{fig:class-distribution}
\end{figure}

\begin{table}[t]
\caption{List of classes along with their definitions.}
\small
\centering
\resizebox{\linewidth}{!}{
\begin{tabular}{|p{1.7cm}|p{6cm}|}
\hline
\textbf{Class}    & \textbf{Definition} \\
\hline\hline
Asphalt    & Road pavement is constructed using aggregates i.e., crushed rocks, sand, and coal tar. \\
\hline
Distress & Longitudinal and transverse cracks occurred due to lack of maintenance. \\
\hline
Gravel     & Unpaved surface with loose aggregation of variable-sized fragments of rocks. \\
\hline
Boggy      & Unpaved road surface filled with mud. \\
\hline
Vegetation Misc   & Naturally occurring vegetation (other than trees) adjacent to the road. \\ % surface growing over the road surface with the                     passage of time. \\
\hline
Crag-stone        &  Hilltop stones---dropped over road surface in mountainous areas.    \\
\hline
Wet Surface       &  Slightly watered road surface; can be damp due to snow or cold weather.  \\
\hline
Road Grime        &  Dirt ingrained on the road.    \\
\hline
Drainage Grate    &  An elongated cover with holes in it or a grating used to cover a water drain.  \\
\hline
Earthen    &  Unpaved roads with compacted layers of stabilized soil. \\
\hline
Water Puddle      &  Small pool of water over the road. \\
\hline
Misc              &  An unclear object dropped over the road. \\
\hline
Concrete   &  Binders such as rough and fine aggregates. \\ %i.e., sand/stones spread over road. \\
\hline
Speed Breaker     &  Concrete speed bumps, speed humps, and speed cushions over the road surface.    \\
\hline
\end{tabular}
}
\label{Table:class-def}
\end{table}

\begin{figure*}[t]
    \centering
    \includegraphics[width=\textwidth]{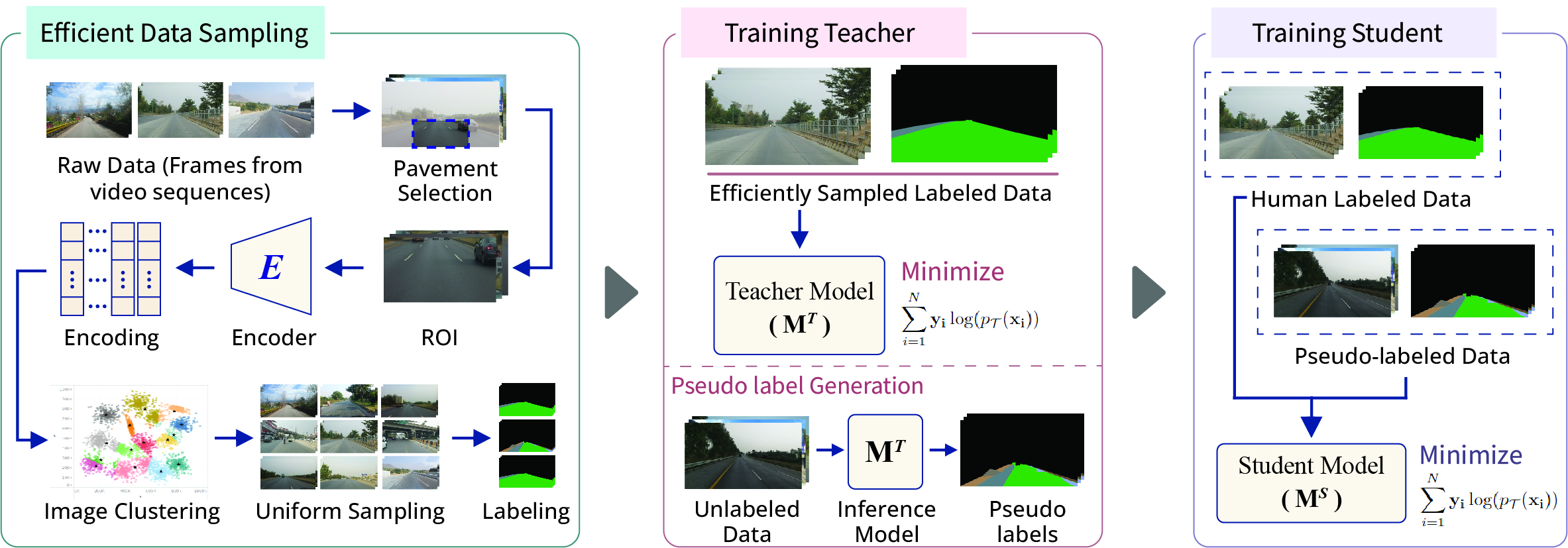}
    \caption{Our \textbf{Efficient Data Sampling (EDS)} based self-training framework. Firstly, raw data samples are clustered based on similarity in road classes among image encodings (shown in Figure~\ref{fig:cluster})---generated by an encoder. Then, a small subset is uniformly formed from all clusters for annotation to train the teacher model. After training, pseudo-labels of the unlabeled set are generated using the teacher model, and the student model is trained on real and pseudo-labeled sets to achieve better generalization.}
    \label{fig:pipeline}
\end{figure*}

\subsection{Training Fully Supervised Baseline Models}
To analyze the effectiveness of R\textsuperscript{2}S100K, we fine-tuned SoTA segmentation networks including FCN \cite{long2015fully}, PSPNet \cite{zhao2017pyramid}, FPN \cite{lin2017feature}, LinkNet \cite{chaurasia2017linknet}, Deeplabv$3$+ \cite{chen2018encoder}, and LRASPP \cite{howard2019searching}, MaskFormer~\cite{cheng2021per}, and SegFromer~\cite{xie2021segformer} along with various backbone networks to perform road segmentation. These methods are trained using a set of human-labeled images ($x, y$) where $x \in R^{H \times W \times 3}$ is a 3-channel RGB image, and $y \in  R^{H \times W \times C}$ is a respective segmentation mask where $H$ and $W$ refers to height and width of the mask, and $C$ indicate classes present in that mask. Following common practices \cite{zhu2019improving}, model $M$ is trained using cross-entropy loss, and IoU is used as a performance metric.
\begin{figure}[t]
    \centering
    \includegraphics[width=\linewidth]{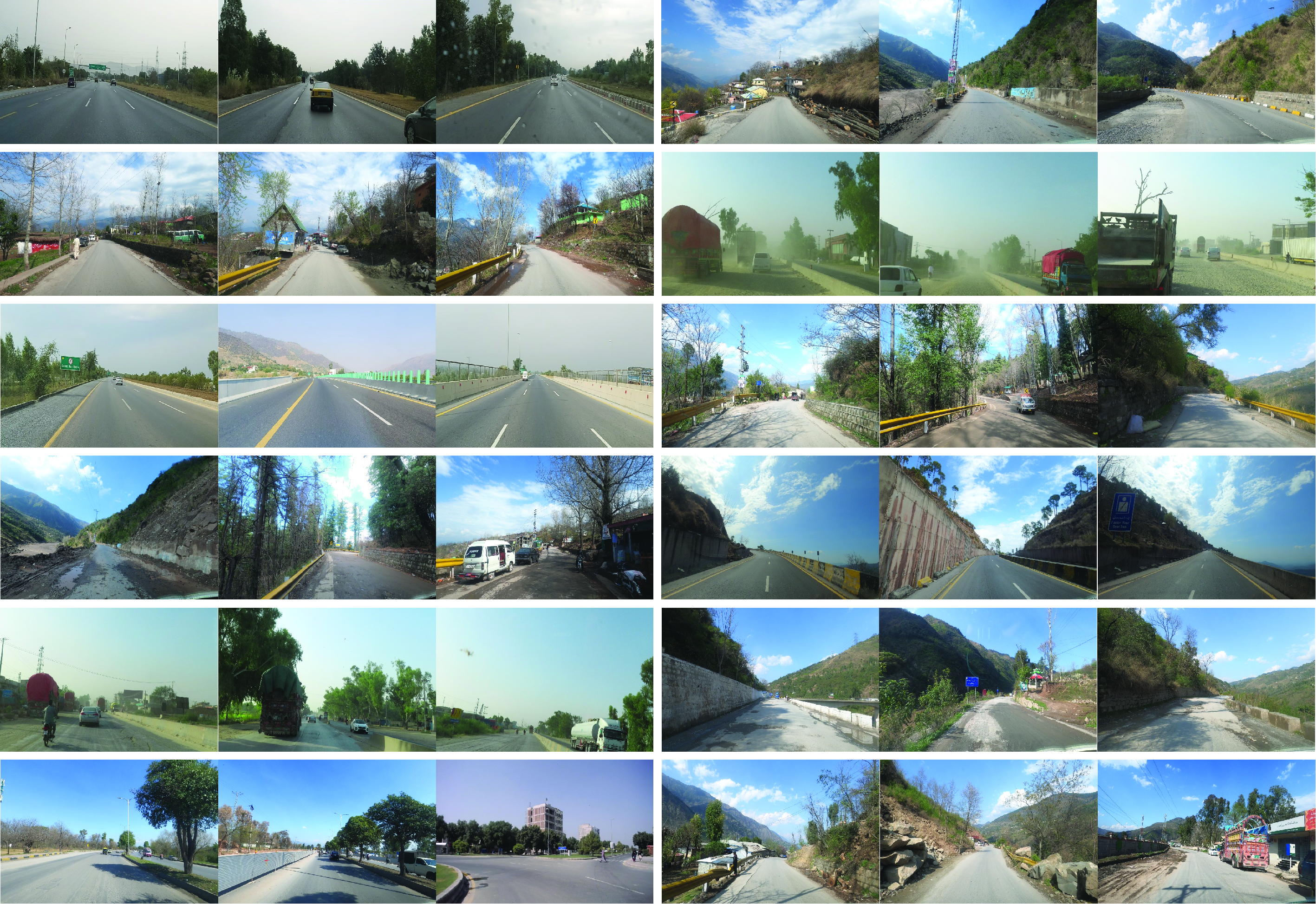}
    \caption{Visualizing examples of clusters (twelve clusters representing three images in each) using our EDS. \textit{Our EDS efficiently clusters images with respect to the similarities in road texture, luminous conditions, and road scenarios.}}
    \label{fig:cluster}
\end{figure}
\subsection{Improving Self-Training Using Unlabeled Data}
Recently, a surge of interest is observed in utilizing unlabeled data to scale up the adaptation of deep models in various segmentation tasks. Leveraging a large amount of unlabeled sets from our R\textsuperscript{2}S100K, we carefully employ semi-supervised training methods to study the generalizability of these models. Taking inspiration from \cite{zhu2019improving}, we employ a teacher-student-based self-training framework to perform road segmentation. Teacher-student-based self-training refers to an approach in which, a large DL model (called teacher) is trained using real labeled data. Then, a set of unlabeled images is given as input to the trained teacher model for inference, and the output of the teacher model is considered as a pseudo-label for the corresponding input image. Finally, data with both---the real and pseudo-labels are combined to train a small/different DL model (called student model) to learn representations from whole data. The purpose of training the teacher model on real data is to guarantee its performance in generating pseudo labels. Therefore, we utilize a small labeled set along with a large unlabeled set to increase the accuracy of the trained model while mitigating the human effort in producing labels at scale.
\subsubsection{Efficient Data Sampling (EDS)}
% In semi-supervised segmentation, dealing with data imbalance problem is a highly challenging task. For scene segmentation, there are mainly two key factors for data imbalance: (i) class-wise pixel ratio, and (ii) class/object confusion. In first case, currently available datasets \cite{cordts2016cityscapes,brostow2009semantic} are quite imbalanced in terms of pixel count for each category due to their long-tail sequential distribution. For instance, a major portion of an image belongs to sky, roads, and roadside environment while the classes like road users i.e., humans, covers far lesser pixel ratio. In second case, it is quite difficult to segment different levels of categories due to class/shape confusion; especially distant objects due to faded representation. Therefore, a teacher model may produce labels with high confidence on major occurring classes, however, its performance may influence in predicting labels with low confidence classes over unlabeled data. Therefore, there is a need for an efficient pipeline to handle data imbalance and improve self-training on segmentation tasks.

In semi-supervised segmentation, dealing with data imbalance problem is a highly challenging task. In street scene segmentation problem, two key factors cause data imbalance; (i) \textit{class imbalance}, which includes class-wise pixel imbalance---a typical image is largely occupied by sky and road, while other classes like humans and bicycles represent far fewer pixels---and class object confusion---some classes, e.g., bicycles, are more challenging to segment due to their complex shapes, occlusions, and faded representations \cite{cordts2016cityscapes, brostow2009semantic}; and (ii) \textit{imbalance in physical scenarios}, as highlighted in Figure \ref{fig:stats}. Although both imbalances are equally important to address, class imbalance is a post-annotation issue that mainly depends on the underlying task, and is generally easily detected, e.g., by computing the confusion matrix of each class. On the contrary, an imbalance in physical scenarios is a pre-annotation issue inherent to the (unlabeled) images themselves. Further, physical scenarios under-represented in the training set are also usually equally under-represented in the test set, and thus, it is significantly more challenging to even detect imbalances in physical scenarios, let alone alleviate them. We identify a dire need for an efficient method to detect/alleviate data imbalances in physical scenarios at the pre-annotation stage to produce more balanced models on semantic segmentation tasks.

To address these issues, we propose EDS, as depicted in Figure \ref{fig:pipeline}. Our goal is to ensure an equal representation of different physical scenarios in the training data. In this regard, our EDS approach has two main stages: (i) data categorization, and (ii) data selection.

\noindent\textbf{Data Categorization:} Firstly, given an unlabeled dataset, $\mathcal{D}_x$, for each $x \in \mathcal{D}_x$, we extract region-of-interest~(ROI) mainly comprising salient road features, sidewalks and pedestrians, while ignoring background, e.g. sky. The extracted image $\text{ROI}(x)$ is then processed through an off-the-shelf encoder network $e(\cdot)$ to get encodings $e(\text{ROI}(x))$. We use a U-Net model, built upon VGG-16 Imagenet encoder, $e: \mathcal{R}^{512\times512\times3} \rightarrow \mathcal{R}^{32\times32\times512}$, as backbone. Due to the prevalent data imbalance problem in segmentation datasets, inherent biases in datasets are also reflected in trained models. Whereas, models trained on Imagenet learn more generic features spanning over 1000 classes, and can be used for multiple downstream tasks. We feel it counter-intuitive to use a biased encoder (trained on street scene dataset) in EDS to mitigate biases in R\textsuperscript{2}S100K.

\begin{figure}[t]
    \centering
    \includegraphics[width=0.9\linewidth]{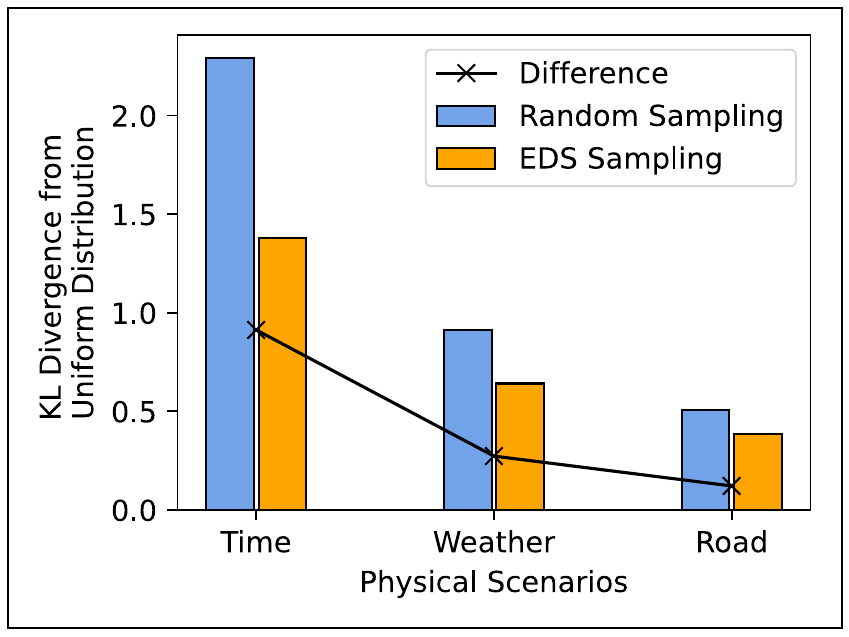}
    \caption{KL divergence between both --- the EDS and Random sampling-based data distributions.}
    \label{fig:stats2}
\end{figure}

\noindent\textbf{Data Selection: } Secondly, encodings $e(\text{ROI}(x))$ of unlabeled train set are fed to $k$-means to get $k$ data clusters $\{C_i\}_{i=1}^{k}$ based on similarities in road surface. Finally, to maintain equal distribution along all types of road representations, we uniformly sample $n$ data instances from each cluster, $C_i$, so that our final dataset, $\mathcal{D}^*_x$ has $n \times k$ data samples. In typical settings, we choose $n \times k = 3000$ to have a comparable dataset size as the cityscapes dataset. Formally,
\begin{equation}
    \mathcal{D}^*_{x} = \cup_{i=1}^k \{ x_j \sim C_i \}_{j=1}^n
\end{equation}
We choose $k=300$, allowing 20 clusters for each class to capture 2(sun/no sun) $\times$ 2(rain/no rain) $\times$ 5(road areas) different scenarios. To compare EDS with random sampling, we sample 500 images from the original dataset using each of the two methods, and compute the probability density of each physical scenario in Figure~\ref{fig:stats2} based on two sampled subsets. Ideally, all labels should have a uniform density, signifying equal representation in the dataset. Therefore, we compute KL-divergence between probability density and uniform distribution in Figure~\ref{fig:stats2}. Results show that EDS significantly improves data imbalance as compared to random sampling.

\subsubsection{Student-Teacher Method For Segmentation Task}
Our self-training framework is illustrated in Figure~\ref{fig:pipeline}. Based on better performance in supervised learning, best-performing model is selected as teacher model $T$ which is used to generate pseudo labels of our unlabeled set of images. The teacher model is used to generate pseudo labels `$y$' of our unlabeled set of images '$x$'. Similar to supervised learning, one-hot encoding of the class labels is sampled from the $p_{\mathcal{T}}(\vb{x})$ as given in equation~\ref{eq:teacher}.

\begin{equation}
    L_{\mathcal{T}} = - \sum_{i=1}^{N} \vb{y_i} \log(p_{\mathcal{T}}(\vb{x_i}))
\label{eq:teacher}
\end{equation}
where, $N$ denotes the number of labeled samples. $y_i$ is the one-hot encoding of class labels, while $p_{\mathcal{T}}$ represents softmax predictions from the teacher model containing class probabilities.

\begin{figure}[t]
    \centering
    \includegraphics[width=\linewidth]{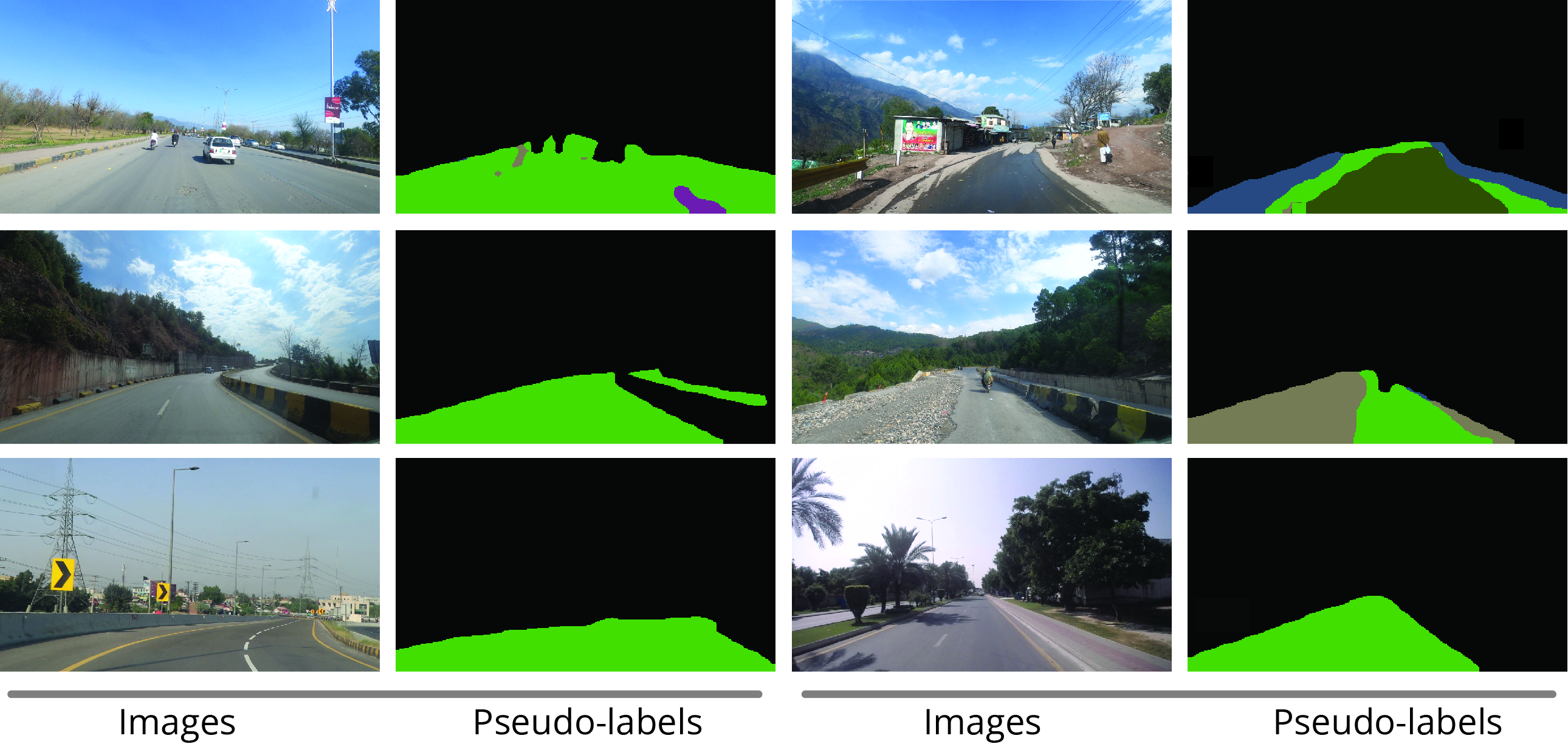}
    \caption{Demonstration of our teacher-generation pseudo-labels over diverse roads. \textit{Our teacher model is able to provide reasonable segmentation predictions.}}
\label{fig:pseudo-labels}
\end{figure}

We demonstrate various examples of our teacher-generated pseudo labels in Figure~\ref{fig:pseudo-labels}. Thanks to our well-performing teacher model, the quality of our teacher-generated pseudo labels $x$ over the unlabeled set is closer to human-annotated labels despite a large domain gap. Therefore, we combine pseudo and real labeled sets to train the student model $S$. Therefore, we combined both the pseudo and real labeled sets to train the student model $S$. Thanks to the generalizability of our proposed self-training pipeline, any DL-based segmentation model can be used as a student model irrespective of their network architectures (briefly explained in section 4.6). Following the practice---adopted in supervised learning, the focus is set to minimize the cross-entropy, given in equation~\ref{eq:student}.

\begin{equation}
    L_{S} = - \sum_{i=1}^{N} \vb{y_i} \log(p_{\mathcal{T}}(\vb{x_i})) - \sum_{j=1}^{M}. \vb{y'_j} \log(p_{\mathcal{S}}(\vb{x'_j}))
\label{eq:student}
\end{equation}
$M$ denotes the number of unlabeled samples. $p_S$ represents softmax predictions from the student model containing the class probabilities. The predicted class probabilities of the student model will be near one-hot by training on hard pseudo-labels generated by the teacher model. Therefore, the entropy of unlabeled data is minimized with cross-entropy loss.

\section{Experiments and Results}
\label{sec:results}

Firstly, we briefly describe the implementation details in terms of hyper-parameter selection for training and evaluation of supervised and semi-supervised learning methods. We categorize our experiments into five sections. In section 4A, we analyze the performance of supervised learning methods and compare the results between random data sampling and our proposed EDS method. In section 4B, we evaluate the performance of semi-supervised learning-based standard self-training methods leveraging our unlabeled data. In section 4C, we select the best-performing semi-supervised model as the teacher method and evaluate its efficacy of the student model with different ratios of unlabeled data samples. In section 4D, we analyze the generalization of other student models irrespective of different network architectures. Lastly, we evaluate the cross-domain generalization with the same categories on state-of-the-art autonomous driving datasets including Cityscapes, CamVid, IDD, and CARL-D.

\subsection{Basic Settings}
To train from scratch, the learning rate is set to 0.002 and 0.0001 for fine-tuning with SGD as an optimizer. As per conventional practice \cite{liu2015parsenet}, a polynomial learning rate is used to smooth learning, and batch size, momentum, and weight decay are set to 8, 0.9, and 0.0001, respectively. Nvidia RTX 3060 is used to perform experiments. The number of training epochs is set to 200 with validation patience of 10 epochs. Evaluation is done using standard Jaccard index (shown in equation \ref{eq:iou}), where FP, TP, and FN refer to the number of false positive, true positive, and false negative pixels, determined over the test set. 

\begin{equation}
    \text{IoU} = \frac{TP}{(TP+FP+TN)}
    \label{eq:iou}
\end{equation}

\subsection{Performance of Supervised Learning with EDS}
We employed FCN \cite{long2015fully}, PSPNet~\cite{zhao2017pyramid}, FPN~\cite{lin2017feature}, LinkNet~\cite{chaurasia2017linknet}, Deeplabv$3$+ \cite{chen2018encoder}, LRASPP \cite{howard2019searching} MaskFormer~\cite{cheng2021per}, and SegFromer~\cite{xie2021segformer} with various backbones on R\textsuperscript{2}S100K. To avoid overfitting, we analyze segmentation methods over a number of labeled subsets (1k, 3k, 5k, 7k, and 9k), randomly sampled from actual 9000 train images. From Table \ref{tab:exp_rs}, it can be seen that employed models---trained over 1k images experience worst performance due to under-fitting. However, their performance significantly improves with 3K train set. Interestingly, the employed models start saturating while training on large train sets, i.e., 5k, 7k, and 9k samples and do not further improve learning because of similarity in road pavement across training samples.

We further analyze the performance of employed models using two data sampling methods i.e., the standard training data selection (STDS) method---in which the data samples are randomly selected based on their occurrence, and our proposed EDS method. It is clear from Table \ref{tab:exp_rs} that segmentation methods perform well with a 3k training set. Therefore, in STDS, we randomly select 3000 labeled images from the train set based on the frequent occurrence. On the other hand, using our EDS, we first clustered all images based on their representation similarities, shown in Fig. \ref{fig:pipeline}. Then, we uniformly sampled out 3000 labeled images from all clusters to form a representative sub-training set.

\begin{table}[t]
    \caption{Evaluation of baseline segmentation methods by training using different numbers of randomly sampled sets from the actual train set of R\textsuperscript{2}S100K dataset for supervised learning.}
    \centering
    \small
    \resizebox{\columnwidth}{!}{
    \begin{tabular}{|l|l|c|c|c|c|c|}
    \hline
    \multirow{2}{*}{Model} & \multirow{2}{*}{Backbone} & \multicolumn{5}{c|}{MIoU} \\ \cline{3-7} 
                       &                           & 1K  & 3K  & 5K & 7K & 9K \\
    \hline\hline
    FCN         &   ResNet-101      &   41.07       & 54.48  & 54.21  & 54.02  & 53.62    \\
    PSPNet      &   ResNet-101     	&   39.83       & 53.03  & 52.96  & 52.43  & 52.14    \\
    LRASPP      &   MobileNet-v3    &   36.31       & 56.54  & 56.19  & 56.10  & 55.93    \\
    FPN         &   ResNet-101      &   44.26       & 55.65  & 54.27  & 54.23  & 54.18    \\
    LinkNet     &   ResNet-101     	&   43.50       & 56.14  & 55.71  & 55.06  & 54.84    \\
    SegFormer   &   -               &   49.77       & 57.86  & 57.60  & 57.48  & 57.21    \\
    MaskFormer  &   -               &   51.35       & 57.98  & 56.37  & 57.72  & 57.05    \\
    \textbf{Deeplab-v3+}  &   \textbf{ResNet-101}      &   45.97       & \textbf{58.02}  & 57.36  & 55.89  & 55.37     \\

    \hline
    \end{tabular}
    }
    \label{tab:exp_rs}
\end{table}

\begin{figure}[t]
\centering
\small
\fbox{\includegraphics[width=\linewidth]{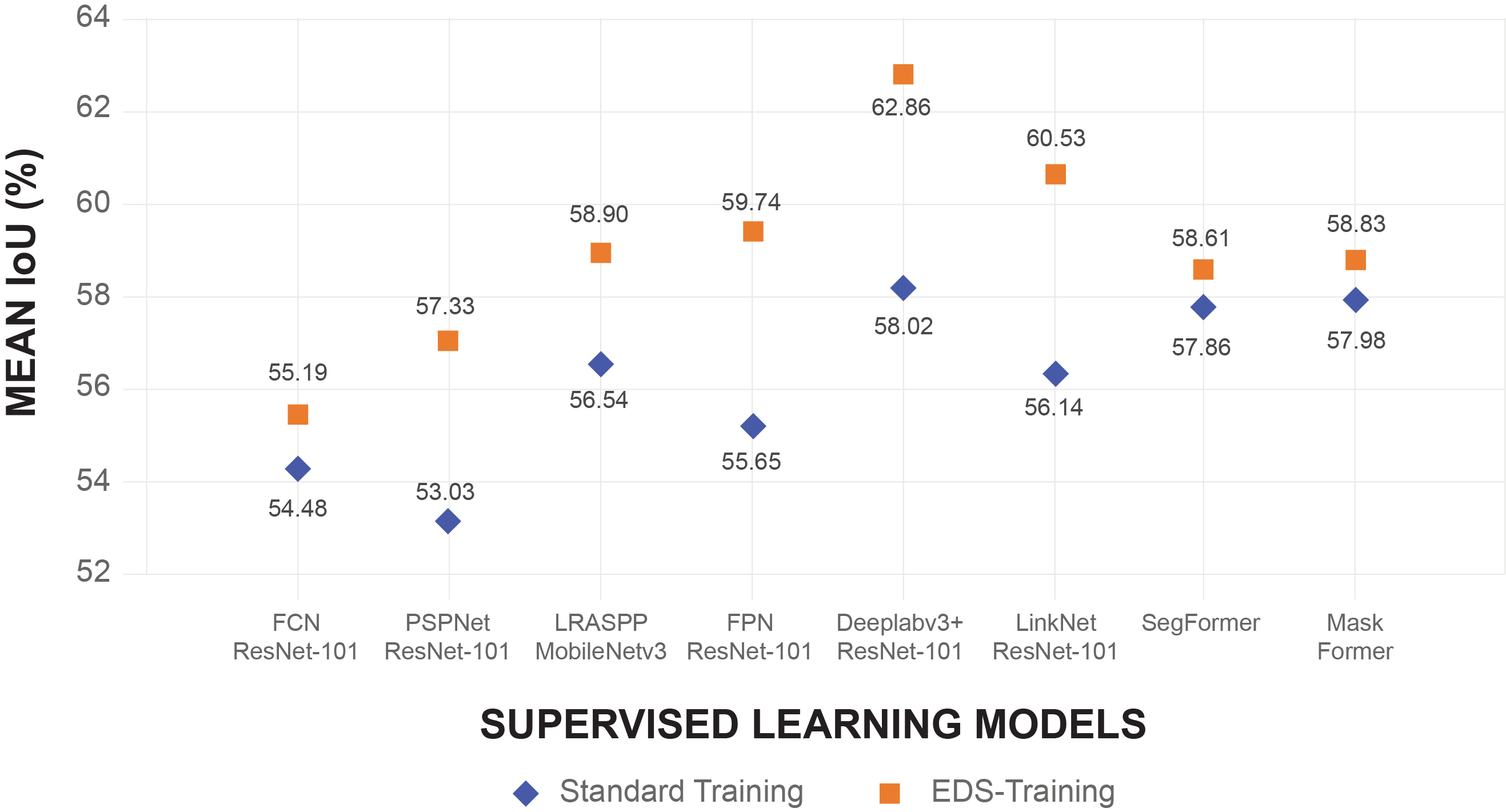}}
\caption{Comparative analysis of baseline segmentation methods using standard data sampling (STDS) and our EDS. \textit{Our efficient data sampling method significantly improves supervised learning for semantic segmentation tasks.}}
\label{fig:Exp1}
\end{figure}

The results illustrated in Figure \ref{fig:Exp1} show that our EDS method significantly improves learning in segmentation tasks. For instance, Deeplab-v3 with ResNet-101 achieved a comparatively highest mIoU, i.e., 62.86\% using our EDS method which is 6.72\% higher than its baseline trained using the STDS method. A major reason for this performance increase is that most of the informative data samples are ignored during random selection, due to which, training data becomes highly unbalanced which ultimately leads to inefficient training and poor generalization. Consequently, the resultant model does not achieve better performance on test data. Whereas, in our EDS method, training samples are uniformly selected based on their class representations. Therefore, the network efficiently learns equal distribution of features from each class that boosts the performance of trained models over test data. A class-wise comparison of state-of-the-art segmentation models is shown in Table \ref{tab:st-exp}.

\begin{table*}[t]
    \caption{Segmentation results (in percentage) of baseline fully-supervised models using EDS on our R\textsuperscript{2}S100K dataset.}
    \centering
    \resizebox{\linewidth}{!}{
    \begin{tabular}{|l|l|c|c|c|c|c|c|c|c|c|c|c|c|c|c|c|}
    \hline
    \multirow{2}{*}{Model}  &  \multirow{2}{*}{Backbone} &   Asphalt  &   Wet  &  Distress & Gravel   &   Boggy    &   Vegetation  &   \multirow{2}{*}{Crag-stone}  &  Road  &  Drainage  &  Earthen  &  Water   &   \multirow{2}{*}{Misc.}    &  Concrete &  Speed & \multirow{2}{*}{MIoU}   \\
    & & Region & Surface & Region & Region & Region & Misc. &  & Grime & Grate & Region & Puddle &  & Region & Breaker & \\
    \hline\hline
    FCN & ResNet-101 & 74.20 & 66.76 & 58.02 & 52.26 & 51.73 & 64.11 & 75.54 & 59.41 & 53.26 & 63.93 & 62.21 & 61.04 & 55.34 & 45.87 & 55.19 \\
    DeeplabV2 & ResNet-101 & 74.31 &  69.59 &	61.48 & 54.27 &	54.51 &	66.72 &	77.10 &	62.37 &	55.51 &	66.45 &	64.92 &	63.26 &	57.53 &	47.15 &	63.15 \\
    FPN & ResNet-101 & 78.08 & 67.58 & 59.74 & 54.32 & 53.87 & 66.79 & 77.25 & 61.10 & 57.77 & 66.90 & 64.40 & 63.88 & 57.17 & 47.46 & 57.33 \\
    FarSeg & ResNeXt-50 & 81.72 & 69.13 & 63.41 & 56.07 & 55.63 & 68.31 & 78.57 & 62.58 & 58.21 & 67.12 & 66.91 & 65.39 & 58.45 & 49.44 & 58.90 \\
    ICNet & ResNeXt-50 & 84.45 & 70.10 & 65.36 & 57.12 & 55.98 & 70.9 & 79.36 & 64.31 & 58.82 & 67.72 & 67.56 & 66.92 & 59.71 & 51.37 & 59.74 \\
    FastSCNN & ResNet-50 & 87.97 & 71.47 & 76.71 & 63.35 & 56.76 & 71.68 & 80.13 & 65.82 & 59.54 & 69.84 & 72.43 & 68.49 & 62.86 & 55.35 & 62.86 \\ 
    HR-Net & ResNeXt-101 & 87.86 & 70.36 & 65.59 & 57.24 & 55.65 & 70.57 & 79.08 & 64.71 & 61.43 & 67.2 & 66.05 & 66.38 & 61.78 & 54.24 & 60.53 \\
    PAN & ResNeXt-101 & 72.02 &	68.87 &	59.11 &	54.2 &	53.76 &	66.41 &	76.13 &	61.27 &	55.47 &	64.19 &	64.47 &	63.51 &	57.91 &	44.58 &	61.56 \\

    \hline
    \end{tabular}
    }
    \label{tab:st-exp}
\end{table*}

\subsection{Effectiveness of Student-Teacher Self-training}
Based on higher performance in supervised learning, we select DeepLabV3+ with ResNet101 as a teacher model to initiate self-training. Firstly, we generate pseudo labels of an unlabeled set with a number of subsets, as shown in Table~\ref{tab:st-ex1}. Then a student model i.e., PSPNet is trained on real and pseudo-labeled sets. From Table~\ref{tab:st-ex1}, it can be observed that utilizing pseudo labels significantly improves segmentation models which is an indication that segmentation models can be improved using pseudo labels without using large-scale labeled data.

\begin{table}[t]
    \caption{Evaluation of EDS-ST on R\textsuperscript{2}S100K.}
    \centering
    \resizebox{0.8\linewidth}{!}{
    \begin{tabular}{|c|c|c|c|c|}
    \hline
    Model   &   Real        &   Pseudo  &   w/ o EDS    &   w EDS \\
    \hline\hline
    Teacher &   3K          & -         &   56.14       &   62.86  \\
    Student &   3K          &  2K       &   59.87       &   63.15\\
    Student  &  3K          &  4K       &   62.24       &   65.82\\
    Student  &  3K          &  8K       &   63.50       &   66.03\\
    Student  &  3K          &  16K      &   62.41       &   66.91\\
    \textbf{Student}  &  \textbf{3K}          &  \textbf{32K}      &   \textbf{62.33}       &   \textbf{67.40} \\
    \hline
    \end{tabular}
    }
    \label{tab:st-ex1}
\end{table}

\subsection{Effectiveness of EDS-based Self-training}
\begin{figure}[t]
    \centering
    \includegraphics[width=\linewidth]{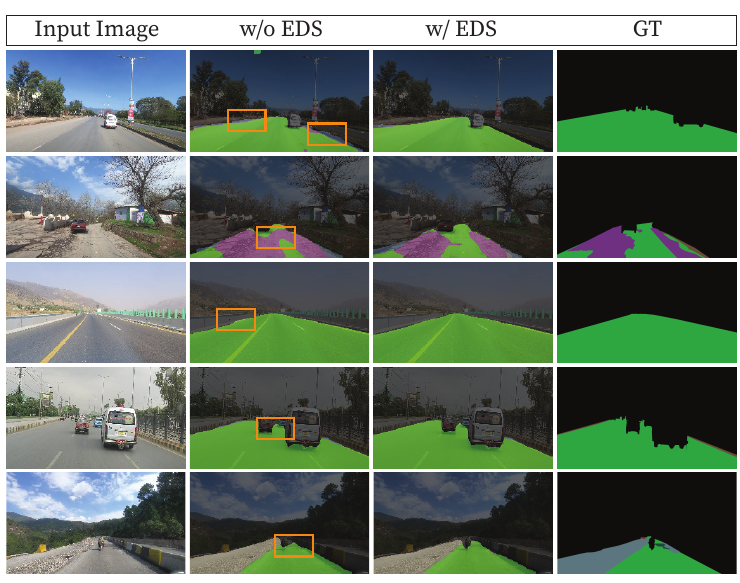}
    \caption{Visualizing the comparison of best-performing student methods on R\textsuperscript{2}S100K. \textit{Results demonstrate that EDS-based self-training is a way better approach to effectively handle class confusion in complex road scenarios.}}
    \label{fig:Exp6}
\end{figure}

Following supervised learning, we used STDS and EDS to analyze the efficient training and its impact on the inference of student models. The results are summarized in Table \ref{tab:st-ex1}, and we have several observations. Firstly, EDS significantly improves student models  with an average increase of 4\% MIoU. Therefore, using EDS for training segmentation models is better than not using it. Secondly, EDS can be used as a generic approach to efficiently train teacher methods. From Figure \ref{fig:Exp1}, it is clear that EDS improves teacher method by 4\%. Thirdly, EDS is necessary to achieve better results when pseudo labels dominate the training set such as the 16k/32k set, otherwise, the performance of the models starts declining. For instance, student models trained without EDS over 16k, and 32k pseudo labeled sets dropped by 0.8\% because of redundant training samples which contribute bias towards classes with more pixels against classes with lesser ones. Whereas EDS efficiently handles data imbalance, thus it improves the performance of student models as compared to the STDS approach, as shown in Figure \ref{fig:Exp6}.

In addition, student models with more pseudo labels (16K, 32K) marginally improve as compared to models with lesser pseudo labels (2K/4K). In the case of fewer pseudo labels, the model learns more informative features as variable data samples are clustered based on similar representation by EDS. However, in the case of more pseudo labels, a vast range of sequential data samples is selected from each cluster, due to which, the model starts saturating instead of learning new information. Whereas, EDS ensures the selection of distinct sampling and helps the model in refining mask boundaries, which is ultimately beneficial for dense tasks.

\subsection{Comparison with related Self-training Methods on R\texorpdfstring{$^2$}{^2}S100K, Cityscapes, and CamVid}
Here we describe a comparative analysis of existing self-training methods. As shown in Table \ref{tab:st_comp}, our EDS outperforms other self-training methods \cite{zou2018unsupervised,abdalla2019fine,zou2019confidence,lee2013pseudo, wang2022semi} on R\textsuperscript{2}100K, as well as on cityscapes and CamVid. On R\textsuperscript{2}100K, consistency regularization achieved 53.70\% mIoU i.e., considerably worse than all of the self-training methods, as the model is learning from inaccurate predictions in the first stage of training, leading to inaccurate inference on test data. Similarly, in the case of teacher fine-tuning, we observe that the model gets stuck at minima at an early stage of fine-tuning. Resultantly, the model starts overfitting instead of learning new information. Similarly, we notice that \cite{wang2022semi} struggles to distinguish hazardous road regions in R\textsuperscript{2}S100K due to higher textural similarities among classes which led to a higher misclassification rate. Whereas, we first efficiently select training data samples using the EDS approach to train a teacher model with considerable accuracy and use it to produce pseudo labels of our unlabeled data. Therefore, its performance consistently improves throughout the training process. Our framework is purely generic; using our approach, a teacher model can train any student model irrespective of their architectural differences which shows its capability of generalization. The performance of EDS is shown in Table~\ref{tab:comparison}.

\begin{table}[t]
    \caption{Evaluation of self-training methods on R\textsuperscript{2}S100K.}
    \centering
    %\small
    \resizebox{\linewidth}{!}{
    \begin{tabular}{|l|l|c|c|} 
    \hline
    Methods                     &   Model    &   MIoU \\
    \hline\hline
    Teacher Fine-tuning \cite{abdalla2019fine}        &  Single Model     &   59.21   \\
    Consistency Regularization \cite{zou2018unsupervised}  &  Single Model   &   53.70   \\
    Model Regularizer \cite{zou2019confidence}           &  Student + Teacher   &   57.64    \\
    Pseudo-labels \cite{lee2013pseudo}              &  Student + Teacher    &   61.05    \\ %\cite{lee2013pseudo}
    U\textsuperscript{2}PL~\cite{wang2022semi}  &  Student + Teacher    & 64.29 \\
    \textbf{EDS (Our Method)}               &  Student + Teacher         &   \textbf{67.40} \\
    \hline
    \end{tabular}
    }
    \label{tab:st_comp}
\end{table}

\begin{table}[t]
  \caption{Analyzing semi-supervised methods on R\textsuperscript{2}S100K.}
  \centering
  \resizebox{\linewidth}{!}{
  \begin{tabular}{|l|l|c|c|c|c|c|c|}
    \hline
    \multirow{2}{*}{Methods} & \multirow{2}{*}{Venue/Year} & \multicolumn{2}{c|}{R\textsuperscript{2}S100K}   &   \multicolumn{2}{c|}{CamVid}          & \multicolumn{2}{c|}{Cityscapes}      \\ \cline{3-8} 
                             &                             & w/o EDS & w EDS & w/o EDS & w EDS  &   w/o EDS    &   w/ EDS \\
    \hline\hline
    
    Baseline \cite{chen2018encoder}     &   CVPR 18     &  62.91    & \textbf{64.27} &  60.82 & \textbf{64.44}    & 62.21 & \textbf{64.73}\\
    CRST \cite{zou2019confidence}       &   ICCV 19     &  63.42    & \textbf{63.79} &  59.37 & \textbf{59.66}    & 60.57 & \textbf{63.86}\\
    HLCon \cite{mittal2019semi}         &   TPAMI 19    &  66.38    & \textbf{66.91} &  62.13 & \textbf{63.71}    & 63.94 & \textbf{64.18}\\
    CCT \cite{ouali2020semi}            &   CVPR 20     &  65.14    & \textbf{65.40} &  63.73 & \textbf{66.10}    & 63.75 & \textbf{65.43}\\
    PseudoSeg \cite{zou2020pseudoseg}   &   ICLR 21     &  64.89    & \textbf{66.73} &  63.58 & \textbf{64.35}    & 64.60 & \textbf{65.98}\\
    %Ours                                &   -           &  & 67.40  \\
    \hline
  \end{tabular}
  }
  \label{tab:comparison}
\end{table}
\subsection{Generalization to Other Student Methods}
Another benefit of EDS-based self-training is that it is not necessary for teacher and student models to have the same architectures. Our framework is a generic pipeline that, firstly, clusters data based on representations; Then, data samples are uniformly selected to ensure data balance for training a teacher model---used to generate pseudo labels which are utilized in improving the accuracy of the student model. In particular, we used DeepLabV3+ with ResNet101 as a teacher model and trained several student models with different backbone networks. These models are selected after analyzing their wide adaptation to segmentation tasks. The results shown in Table \ref{tab:st_gen} demonstrate that EDS-based self-training can significantly improve student models irrespective of their architectures. Comparatively, PSPNet with ResNet101 outperformed other segmentation networks by using the EDS approach.

\begin{table}[t]
    \caption{Generalizability of student methods irrespective of different backbone network architectures on R\textsuperscript{2}100K.}
    \centering
    \resizebox{\linewidth}{!}{
    \begin{tabular}{|l|l|c|c|c|c|}
    \hline
    Model       &   Backbone    &   Val mIoU &   Test mIoU \\
    \hline\hline
    BiSeNet         &   ResNet-50       &   62.32   &	63.17    \\
    BiSeNet w/ EDS         &   ResNet-50      & 64.40	&	64.93   \\

    PSPNet         &   ResNet-101     &   62.77	&	64.51   \\
    PSPNet w/ EDS        &   ResNet-101    &     65.82	&	67.23  \\

    LRASPP      &   MobileNet-v3  &  59.11	&	59.87  \\
    LRASPP w/ EDS     &   MobileNet-v3    &     60.56	&	61.38  \\
    
    LinkNet   &   ResNet-101    &       62.48	&	63.59  \\
    LinkNet w/ EDS  &   ResNet-101     &     64.25	&	64.72  \\
    \hline
    \end{tabular}
    }
    \label{tab:st_gen}
\end{table}

\section{Conclusions}
\label{sec:cons}
In this paper, we presented R\textsuperscript{2}S100K to perform drivable road region segmentation on unstructured roadways. Alongside, we presented a self-training framework to improve semi-supervised learning for segmentation tasks. Results demonstrate that our proposed method can be utilized to improve supervised/semi-supervised learning for semantic segmentation due to its effective class confusion handling in complex road environments. We believe that our training framework will facilitate research in various ML applications, where generating labeled data is a critical task.

\bibliographystyle{IEEEtran}
% \bibliography{refs}

\end{document}